\documentclass[10pt,twocolumn,letterpaper]{article}

\usepackage{iccv}
\usepackage{times}
\usepackage{epsfig}
\usepackage{graphicx}
\usepackage{amsmath}
\usepackage{amssymb}
\usepackage{bm}
\usepackage{algorithm}

\newtheorem{theo}{Theorem}
\newtheorem{lem}{Lemma}

\usepackage[breaklinks=true,bookmarks=false]{hyperref}

\iccvfinalcopy 

\def\iccvPaperID{****} 

\ificcvfinal\fi
\begin{document}

\title{Correlation Adaptive Subspace Segmentation by Trace Lasso}

\author{Canyi Lu$^1$, Jiashi Feng$^1$, Zhouchen Lin$^{2,}$\thanks{Corresponding author.},  \ Shuicheng Yan$^1$\\
$^1$ Department of Electrical and Computer Engineering, National University of Singapore\\
$^2$ Key Laboratory of Machine Perception (MOE), School of EECS, Peking University\\
{\tt\small canyilu@gmail.com, a0066331@nus.edu.sg, zlin@pku.edu.cn, eleyans@nus.edu.sg}
}

\maketitle
\thispagestyle{empty}
\begin{abstract}

This paper studies the subspace segmentation problem. Given a set of data points drawn from a union of subspaces, the goal is to partition them into their underlying subspaces they were drawn
from. The spectral clustering method is used as the framework. It requires to find an affinity matrix which is close to block diagonal, with nonzero entries corresponding to the data point
pairs from the same subspace. In this work, we argue that both sparsity and the grouping effect are important for subspace segmentation. A sparse affinity matrix tends to be block diagonal,
with less connections between data points from different subspaces. The grouping effect ensures that the highly corrected data which are usually from the same subspace can be grouped
together. Sparse Subspace Clustering (SSC), by using $\ell^1$-minimization, encourages sparsity for data selection, but it lacks of the grouping effect. On the contrary, Low-Rank
Representation (LRR), by rank minimization, and Least Squares Regression (LSR), by $\ell^2$-regularization, exhibit strong grouping effect, but they are short in subset selection. Thus the
obtained affinity matrix is usually very sparse by SSC, yet very dense by LRR and LSR.

In this work, we propose the Correlation Adaptive Subspace Segmentation (CASS) method by using trace Lasso. CASS is a data correlation dependent method which simultaneously performs automatic data selection and groups correlated data together. It can be regarded as a method which adaptively balances SSC and LSR. Both theoretical and experimental results show the effectiveness of CASS.



\end{abstract}

\section{Introduction}
This paper focuses on subspace segmentation, the goal of which is to segment a given data set into clusters, ideally with each cluster corresponding to a subspace. Subspace segmentation is an important problem in both computer vision and machine learning literature. It has numerous applications, such as motion segmentation \cite{Motion2010PAMI}, face clustering \cite{LRRpami}, and image segmentation \cite{facecluster}, owing to the fact that the real-world data often approximately lie in a mixture of subspaces. The problem is formally defined as follows \cite{LRR}:

\newtheorem{definition}{Definition}
\begin{definition}
(Subspace Segmentation) Given a set of sufficiently sampled data vectors $X=[x_1,\cdots,x_n]\in \mathbb{R}^{d\times n}$, where $d$ is the feature dimension, and $n$ is the number of data vectors. Assume that the data are drawn from a union of $k$ subspaces $\{\mathcal{S}_i\}_{i=1}^k$ of unknown dimensions $\{r_i\}_{i=1}^k$, respectively. The task is to segment the data according to the underlying subspaces they are drawn from.
\end{definition}

\begin{figure}[!t]
\centering
\includegraphics[width=0.5\textwidth]{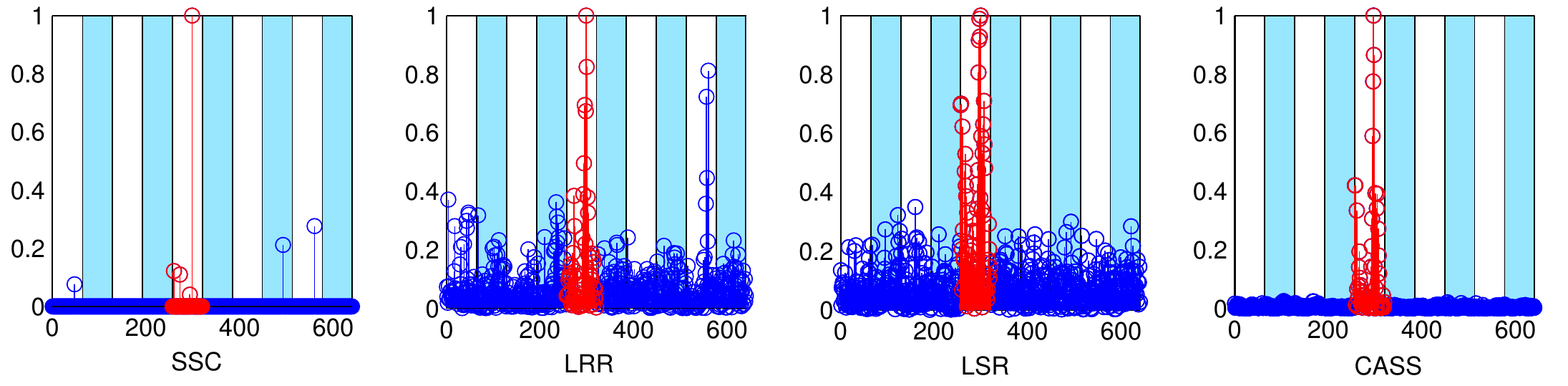}
\caption{\small{\textbf{Example on a subset with 10 subjects of
the Extended Yale B database.} For a given data point $y$ and a
data set $X$, $y$ can be approximately expressed as a liner
representation of all the columns of $X$ by different methods.
This figure shows the absolute values of the representation
coefficients (normalized to [0 1] for ease of display) derived by
SSC, LRR, LSR and the proposed CASS. Here different columns in
each subfigure indicate different subjects. The red color
coefficients correspond to the face images which are from the same
subject as $y$. One can see that the coefficients derived by SSC
are very sparse, and only limited samples within cluster are
selected to represent $y$. Both LRR and LSR lead to dense
representations. They not only group data within cluster together,
but also between clusters. For CASS, most of large coefficients
concentrate on the data points within cluster. Thus it
approximately reveals the true segmentation of data. {\bf Images
in this paper are best viewed on screen!}}} \label{fig_grouping}
\end{figure}
\subsection{Summary of notations}
Some notations are used in this work. We use capital and lowercase
symbols to represent matrices and vectors, respectively. In
particular, $\bm{1}_d\in\mathbb{R}^d$ denotes the vector of all
1's, $e_i$ is a vector whose $i$-th entry is 1 and 0 for others,
and $I$ is used to denote the identity matrix. Diag($v$) converts
the vector $v$ into a diagonal matrix in which the $i$-th diagonal
entry is $v_i$. diag($A$) is a vector whose $i$-th entry is
$A_{ii}$ of a square matrix $A$. tr($A$) is the trace of a square
matrix $A$. $A_i$ denotes the $i$-th column of a matrix $A$.
sign($x$) is the sign function defined as sign$(x)=x/|x|$ if
$x\neq0$ and 0 for otherwise. $v\rightarrow v_0$ denotes that $v$
converges to $v_0$.

Some vector and matrix norms will be used. $||v||_0$, $||v||_1$,
$||v||_2$ and $||v||_\infty$ denote the $\ell^0$-norm (number of
nonzero entries),  $\ell^1$-norm (sum of the absolute vale of each
entry), $\ell^2$-norm and $\ell^\infty$-norm of a vector $v$.
$||A||_1$, $||A||_F$, $||A||_{2,1}$, $||A||_\infty$, and $||A||_*$
denote the $\ell^1$-norm ($\sum_{i,j}|A_{ij}|$), Frobenius norm,
$\ell^{2,1}$-norm ($\sum_j||A_j||_2$),
$\ell^\infty$-norm ($\max_{i,j}|A_{ij}|$), and nuclear norm (the
sum of all the singular values) of a matrix $A$, respectively.

\subsection{Related work}
There has been a large body of research on subspace segmentation
\cite{GPCA,SSC,LRR,SSQP,MSR,fang2012graph,gui2012discriminant}. Most
recently, the Sparse Subspace Clustering (SSC)
\cite{SSC,elhamifar2012sparse}, Low-Rank Representation (LRR)
\cite{LRR,LRRpami,L1Graph}, and Least Squares Regression (LSR)
\cite{LSR} techniques have been proposed for subspace segmentation
and attracted much attention. These methods learn an affinity
matrix whose entries measure the similarities among the data
points and then perform spectral clustering on the affinity matrix
to segment data. Ideally, the affinity matrix should be block
diagonal (or block sparse in vector form), with nonzero entries
corresponding to data point pairs from the same subspace. A
typical choice for the measure of similarity between $x_i$ and
$x_j$ is $W_{ij}=\exp{(-||x_i-x_j||/\sigma)}$, where $\sigma>0$. However, such method is unable to utilize the
underlying linear subspace structure of data. The constructed
affinity matrix is usually not block diagonal even under certain
strong assumptions, \emph{e.g.} independent subspaces  \footnote{A
collection of $k$ linear subspaces $\{\mathcal{S}_i\}_{i=1}^k$ are
independent if and only if $\mathcal{S}_i\cap\sum_{j\neq
i}\mathcal{S}_j=\{0\}$ for all $i$ (or
$\sum_{i=1}^k\mathcal{S}_i=\oplus_{i=1}^k\mathcal{S}_i$).}. For a
new point $y\in\mathbb{R}^d$ in the subspaces, SSC pursues a
sparse representation:
\begin{equation}\label{SRclean}
    \min_{w} ||w||_1 \ \ \text{s.t.} \  y=Xw.
\end{equation}
Problem (\ref{SRclean}) can be extended for handling the data with noise, which leads to the popular Lasso \cite{Lasso} formulation:
\begin{equation}
\label{SR}
\min_{w} \ ||y-Xw||_2^2+\lambda||w||_1,
\end{equation}
where $\lambda>0$ is a parameter. SSC solves problem (\ref{SRclean}) or (\ref{SR})
for each data point $y$ in the dataset with all the other data
points as the dictionary. Then it uses the derived representation
coefficients to measure the similarities between data points and
constructs the affinity matrix. It is shown that, if the subspaces
are independent, the sparse representation is block sparse.
However, if the data from the same subspace are highly correlated
or clustered, the $\ell^1$-minimization will generally select a
single representative at random, and ignore other correlated data.
This leads to a sparse solution but misses data correlation
information. Thus SSC may result in a sparse affinity
matrix but lead to unsatisfactory performance.

Low-Rank Representation (LRR) is a method which aims to group the correlated data together. It solves the following convex optimization problem:
\begin{equation}\label{LRRclean}
\min_{W}        \   ||W||_*   \ \  \text{s.t.} \   X=XW.
\end{equation}
The above problem can be extended for the noisy case:
\begin{equation}\label{LRR}
    \begin{split}
    \min_{W,E}  \  & ||W||_* + \lambda ||E||_{2,1}    \\
    \text{s.t.} \  & X=XW+E,
    \end{split}
\end{equation}
where $\lambda>0$ is a parameter. Although LRR guarantees
to produce a block diagonal solution when the data are noise free
and drawn from independent subspaces, the real data are usually
contaminated with noises or outliers. So the solution to problem
(\ref{LRR}) is usually very dense and far from block diagonal. The
reason is that the nuclear norm minimization lacks the ability of
subset selection. Thus, LRR generally groups correlated data
together, but sparsity cannot be achieved.

In the context of statistics, Ridge regression
($\ell^2$-regularization) \cite{Ridge} may have the similar
behavior as LRR. Below is the most recent work by using Least
Squares Regression (LSR) \cite{LSR} for subspace segmentation:

\begin{equation}\label{LSR}
    \min_{W}  \  ||X-XW||_F^2 + \lambda ||W||_F^2.
\end{equation}
Both LRR and LSR encourage grouping effect but lack of sparsity.
In fact, for subspace segmentation, both sparsity and grouping
effect are very important. Ideally, the affinity matrix should be
sparse, with no connection between clusters. On the other hand,
the affinity matrix should not be too sparse, \emph{i.e.}, the
nonzero connections within cluster should be sufficient enough for
grouping correlated data in the same subspaces. Thus, it is
expected that the model can automatically group the correlated
data within cluster (like LRR and LSR) and eliminate the
connections between clusters (like SSC). Trace Lasso
\cite{TraceLasso}, defined as $||X\text{Diag}(w)||_*$, is such a
newly established regularizer which interpolates between the
$\ell^1$-norm and $\ell^2$-norm of $w$. It is \emph{adaptive} and
depends on the correlation among the samples in $X$, which can be
encoded by $X^TX$. In particular, when the data are highly
correlated ($X^TX$ is close to $\bm{1}\bm{1}^T$), it will be close
to the $\ell^2$-norm, while when the data are almost uncorrelated
($X^TX$ is close to $I$), it will behave like the $\ell^1$-norm.
We take the \emph{adaptive} advantage of trace Lasso to regularize
the representation coefficient matrix, and define an affinity
matrix by applying spectral clustering to the normalized
Laplacian. Such a model is called Correlation Adaptive Subspace
Segmentation (CASS) in this work. CASS can be regarded as a method
which adaptively interpolates SSC and LSR. An intuitive comparison
of the coefficient matrices derived by these four methods can be
found in Figure \ref{fig_grouping}. For CASS, we can see that most
large representation coefficients cluster on the data points from
the same subspace as $y$. In comparison, the connections within
cluster are very sparse by SSC, and the connections between
clusters are very dense by LRR and LSR.

\subsection{Contributions}
We summarize the contributions of this paper as follows:
\begin{itemize}
  \item We propose a new subspace segmentation method, called the Correlation Adaptive Subspace Segmentation (CASS), by using trace Lasso \cite{TraceLasso}. CASS is the first method that takes the data correlation into account for subspace segmentation. So it is self-adaptive for different types of data.
  \item In theory, we show that if the data are from independent subspaces, and the objective function satisfies the proposed Enforced Block Sparse (EBS) conditions, then the obtained solution is block sparse. Trace Lasso is a special case which satisfies the EBS conditions.
  \item We theoretically prove that trace Lasso has the grouping effect, \emph{i.e.}, the coefficients of a group of correlated data are approximately equal.
\end{itemize}

\section{Correlation Adaptive Subspace Segmentation by Trace Lasso}

Trace Lasso \cite{TraceLasso} is a recently proposed norm which
balances the $\ell^1$-norm and $\ell^2$-norm. It is formally
defined as
\begin{equation*}
\Omega(w)=||X\text{Diag}(w)||_*.
\end{equation*}
A main difference between trace Lasso and the existing norms is
that trace Lasso involves the data matrix $X$, which makes it
\emph{adaptive} to the correlation of data. Actually, it only
depends on the matrix $X^TX$ of data, which encodes the
correlation information among data. In particular, if the norm of
each column of $X$ is normalized to one, we have the following
decomposition of $X\text{Diag}(w)$:
\begin{equation*}
    X\text{Diag}(w) = \sum_{i=1}^n |w_i|(\text{sign}(w_i)x_i)e_i^T.
\end{equation*}
If the data are uncorrelated (the data points are orthogonal, $X^TX=I$), the above equation gives the singular value decomposition of $X\text{Diag}(w)$. In this case, trace Lasso is equal to the $\ell^1$-norm:
\begin{equation*}
    ||X\text{Diag}(w)||_*= ||\text{Diag}(w)||_* = \sum_{i=1}^n|w_i|=||w||_1.
\end{equation*}
If the data are highly correlated (the data points are all the same, $X=x_1\bm{1}^T$, $X^TX=\bm{1}\bm{1}^T$), trace Lasso is equal to the $\ell^2$-norm:
\begin{equation*}
    ||X\text{Diag}(w)||_*=||x_1w^T||_*=||x_1||_2||w||_2=||w||_2.
\end{equation*}
For other cases, trace Lasso interpolates between the $\ell^2$-norm and $\ell^1$-norm \cite{TraceLasso}:
\begin{equation*}
    ||w||_2\leq||X\text{Diag}(w)||_*\leq ||w||_1.
\end{equation*}

We use trace Lasso for subset selection from all the data adaptively, which leads to the Correlation Adaptive Subspace Segmentation (CASS) method. We first consider the subspace segmentation problem with clean data by CASS and then extend it to the noisy case.

\subsection{CASS with clean data}

Let $X=[x_1,\cdots,x_n]=[X_1,\cdots,X_k]\Gamma$ be a set of data
drawn from $k$ subspaces $\{\mathcal{S}_i\}_{i=1}^k$, where $X_i$
denotes a collection of $n_i$ data points from the $i$-th subspace
$\mathcal{S}_i$, $n=\sum_{i=1}^{k}n_i$, and $\Gamma$ is a
hypothesized permutation matrix which rearranges the data to the
true segmentation of data. For a given data point
$y\in\mathcal{S}_i$, it can be represented as a linear combination
of all the data points $X$. Different from the previous methods in
SSC, LRR and LSR, CASS uses the trace Lasso as the objective
function and solves the following problem:
\begin{equation}\label{CASS}
    \min_{w\in\mathbb{R}^n}  ||X\text{Diag}(w)||_*  \ \ \text{s.t.}  \ \ y=Xw.
\end{equation}

The methods, SSC, LRR and LSR, show that if the data are
sufficiently sampled from independent subspaces, a block diagonal
solution can be achieved. The work \cite{LSR} further shows that
it is easy to get a block diagonal solution if the objective
function satisfies the Enforced Block Diagonal (EBD) conditions.
But the EBD conditions cannot be applied to trace Lasso directly,
since trace Lasso is a function involving both the data $X$ and
$w$. Here we extend the EBD conditions \cite{LSR} to the Enforced
Block Sparse (EBS) conditions and show that the obtained solution
is block sparse when the objective function satisfies the EBS
conditions. Trace Lasso is a special case which satisfies the EBS
conditions and thus leads to a block sparse solution.

\textbf{Enforced Block Sparse (EBS) Conditions.} Assume $f$ is a
function with regard to a matrix $X\in\mathbb{R}^{d\times n}$ and
a vector $w=[w_a;w_b;w_c]\in\mathbb{R}^n$, $w\neq 0$. Let
$w^B=[0;w_b;0]\in\mathbb{R}^n$. The EBS conditions are:
\begin{enumerate}
\renewcommand{\labelenumi}{(\theenumi)}
\item $f(X,w)=f(XP,P^{-1}w)$, for any permutation matrix $P\in\mathbb{R}^{n\times n}$;
\item $f(X,w)\geq f(X,w^B)$, and the equality holds if and only if $w=w^B$.
\end{enumerate}
For some cases, the EBS conditions can be regarded as extensions
of the EBD conditions \footnote{For example,
$f(X,w)=||w||_p+0\times ||X||_F=||w||_p=g(w)$, where $p\geq0$. It is
easy to see that $f(X,w)$ satisfies the EBS conditions and $g(w)$
satisfies the EBD conditions.}. The EBS conditions will enforce
the solution to the following problem
\begin{equation}\label{EBSgener}
    \min_w f(X,w)  \ \  \text{s.t.} \ \  y=Xw,
\end{equation}
to be block sparse when the subspace are independent.
\begin{theo}
\label{ThmBlock1} Let
$X=[x_1,\cdots,x_n]=[X_1,\cdots,X_k]\Gamma\in\mathbb{R}^{d\times
n}$ be a data matrix whose column vectors are sufficiently
\footnote{That the data sampling is sufficient makes sure that
problem (\ref{EBSgener}) has a feasible solution.} drawn from a
union of $k$ independent subspaces $\{\mathcal{S}_i\}_{i=1}^k$,
$x_j\neq0$, $j=1,\cdots,n$. For each $i$,
$X_i\in\mathbb{R}^{d\times n_i}$ and $n=\sum_{i=1}^k n_i$.  Let
$y\in\mathbb{R}^d$ be a new point in $\mathcal{S}_i$. Then the
solution to problem (\ref{EBSgener})
$w^*=\Gamma^{-1}[z_1^*;\cdots;z_k^*]\in\mathbb{R}^n$ is block
sparse, \emph{i.e.}, $z_i^*\neq 0$ and $z_j^*=0$ for all $j\neq
i$.
\end{theo}
\emph{Proof.} For $y\in\mathcal{S}_i$, let $w^*=\Gamma^{-1}[z_1^*;\cdots;z_k^*]$ be the optimal solution to problem (\ref{EBSgener}), where $z_i^*\in\mathbb{R}^{n_i}$ corresponds to $X_i$ for each $i=1,\cdots,k$. We decompose $w^*$ into two parts $w^*=u^*+v^*$, where $u^*=\Gamma^{-1}[0;\cdots;z_i^*;\cdots;0]$ and $v^*=\Gamma^{-1}[z_1^*;\cdots;0;\cdots;z_k^*]$. We have
\begin{equation*}
    \begin{split}
    y & = Xw^* = Xu^* + Xv^* \\
      & = X_iz_i^* + \sum_{j\neq i}X_jz_j^*.  \\
    \end{split}
\end{equation*}
Since $y\in\mathcal{S}_i$ and $X_iz_i^*\in\mathcal{S}_i$,
$y-X_iz_i^*\in\mathcal{S}_i$. Thus $\sum_{j\neq
i}X_jz_j^*=y-X_iz_i^*\in\mathcal{S}_i\cap\oplus_{j\neq
i}\mathcal{S}_j$. Considering that the subspaces
$\{\mathcal{S}_i\}_{i=1}^k$ are independent,
$\mathcal{S}_i\cap\oplus_{j\neq i}\mathcal{S}_j=\{0\}$, we have
$y=X_iz_i^*=Xu^*$ and $X_jz_j^*=0$, $j\neq i$. So $u^*$ is
feasible to problem (\ref{EBSgener}). On the other hand, by the
definition of $u^*$ and the EBS conditions (2), we have
\begin{equation*}
    f(X,w^*)\geq f(X,u^*).
\end{equation*}
Noticing that $w^*$ is optimal to problem (\ref{EBSgener}),
$f(X,w^*)\leq f(X,u^*)$. Thus the equality holds. By the EBS
conditions (2), we get $w^*=u^*$. Therefore, $z_i^*\neq0$, and
$z_j^*=0$ for all $j\neq i$. $\hfill\blacksquare$

The EBS conditions greatly extend the family of the objective function which involves the block sparse property. It is easy to check that trace Lasso satisfies the EBS conditions. Let $f(X,w)=||X\text{Diag}(w)||_*$, for any permutation matrix $P\in\mathbb{R}^{n\times n}$,
\begin{equation*}
    \begin{split}
     f(XP,P^{-1}w) = & ||XP\text{Diag}(P^{-1}w)||_* \\
    = & ||XPP^{-1}\text{Diag}(w)||_* \\
    = & ||X\text{Diag}(w)||_* = f(X,w).
    \end{split}
\end{equation*}
Trace Lasso also satisfies the EBS conditions (2) by the following lemma:
\begin{lem}
\label{Lemma} \cite[Lemma 11]{LRRsemi} Let
$A\in\mathbb{R}^{d\times n}$ be partitioned in the form
$A=[A_1,A_2]$. Then $||A||_*\geq||A_1||_*$ and the equality holds
if and only if $A_2=0$.
\end{lem}

In a similar way, CASS owns the block sparse property:
\begin{theo}
\label{ThmBlock}
Let $X=[x_1,\cdots,x_n]=[X_1,\cdots,X_k]\Gamma\in\mathbb{R}^{d\times n}$ be a data matrix whose column vectors are sufficiently drawn from a union of $k$ independent subspaces $\{\mathcal{S}_i\}_{i=1}^k$, $x_j\neq0$, $j=1,\cdots,n$. For each $i$, $X_i\in\mathbb{R}^{d\times n_i}$ and $n=\sum_{i=1}^k n_i$.  Let $y$ be a new point in $\mathcal{S}_i$. It holds that the solution to problem (\ref{CASS}) $w^*=\Gamma^{-1}[z_1^*;\cdots;z_k^*]\in\mathbb{R}^n$ is block sparse, \emph{i.e.}, $z_i^*\neq 0$ and $z_j^*=0$ for all $j\neq i$. Furthermore, $z_i^*$ is also optimal to the following problem:
\begin{equation}\label{CASS_sub}
        \min_{z_i\in\mathbb{R}^{n_i}}   ||X_i\text{Diag}(z_i)||_*  \ \ \   \text{s.t.}  \   \  y=X_iz_i.
\end{equation}
\end{theo}

The block sparse property of CASS is the same as those of SSC, LRR
and LSR when the data are from independent subspaces. This is also
the motivation for using trace Lasso for subspace segmentation.
For the noisy case, different from the previous methods, CASS may
also lead to a solution which is close to block sparse, and it
also has the grouping effect (see Section \ref{sec_grouping}).

\subsection{CASS with noisy data}
The noise free and independent subspaces assumption may be
violated in real applications. Problem (\ref{CASS}) can be
extended to handle noises of different types. For small magnitude
and dense noises (\emph{e.g.} Gaussian), a reasonable strategy is
to use the $\ell^2$-norm to model the noises:
\begin{equation}\label{CASSl2}
    \min_{w}        \  \  \frac{1}{2}||y-Xw||_2^2 + \lambda||X\text{Diag}(w)||_*.
\end{equation}
Here $\lambda>0$ is a parameter balancing the effects of the two terms. For data with a small fraction of gross corruptions, the $\ell^1$-norm is a better choice:
\begin{equation}\label{CASSl1}
    \min_{w}        \  \ ||y-Xw||_1 + \lambda||X\text{Diag}(w)||_*.
\end{equation}
Namely, the choice of the norm depends on the noises. It is important for subspace segmentation but not the main focus of this paper.
\begin{figure}[!t]
\centering
\includegraphics[width=0.5\textwidth]{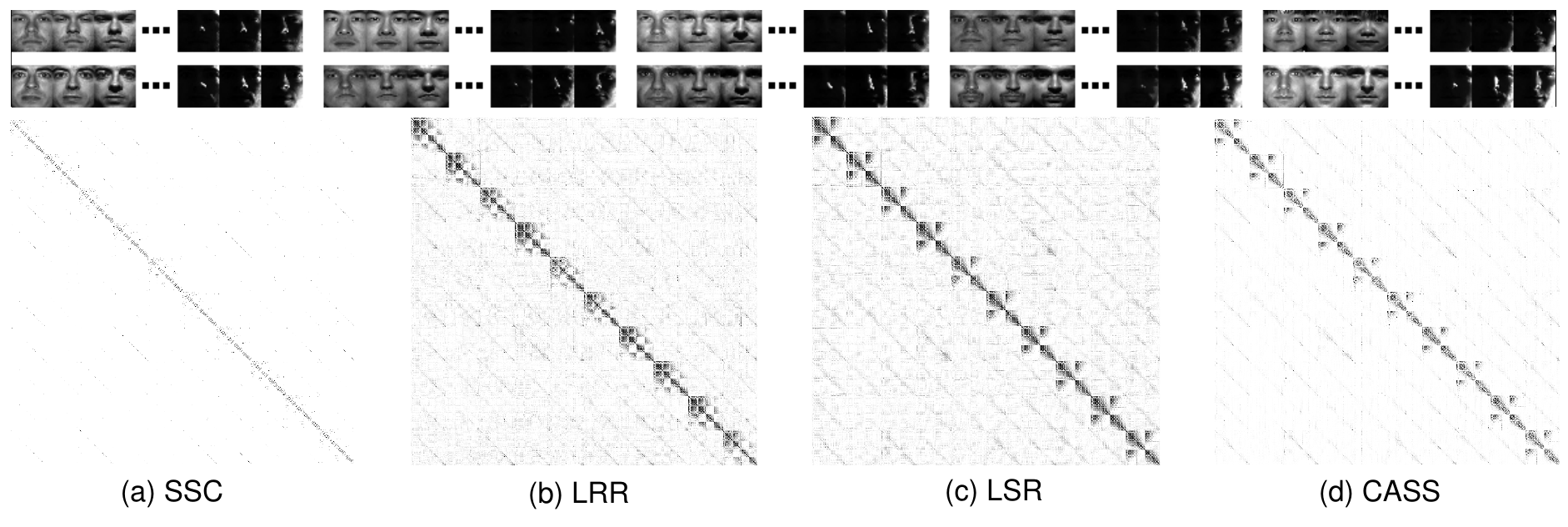}
\caption{\small{The affinity matrices derived by (a) SSC, (b) LRR, (c) LSR, and (d) CASS on the Extended Yale B Database (10 subjects).}}
\label{fig_YaleB_graphs}
\end{figure}

In the case of data contaminated with noises, it is difficult to
obtain a block sparse solution. Though the representation
coefficient derived by SSC tends to be sparse, it is unable to
group correlated data together. On the other hand, LRR and LSR
lead to dense representations which lack  the ability of subset
selection. CASS by using trace Lasso takes the correlation of data
into account which places a tradeoff between sparsity and grouping
effect. Thus it can be regarded as a method which balances SSC and
LSR.

For SSC, LRR, LSR and CASS, each data point is expressed as a
linear combination of all the data with a coefficient vector.
These coefficient vectors can be arranged as a matrix measuring
the similarities between data points. Figure
\ref{fig_YaleB_graphs} illustrates the coefficient matrices
derived by these four methods on the Extended Yale B database (see
Section \ref{subsec_exp} for detailed experimental setting). We
can see that the coefficient matrix derived by SSC is so sparse
that it is even difficult to identify how many groups there are.
This phenomenon confirms that SSC loses the data correlation
information. Thus SSC does not perform well for data with strong
correlation. On the contrary, the coefficient matrices derived by
LRR and LSR are very dense. They group many data points together,
but do not do subset selection. There are many nonzero connections
between clusters, and some are very large. Thus LRR and LSR may
contain much erroneous information. Our proposed method CASS by
using trace Lasso, achieves a more accurate coefficient matrix,
which is close to be block diagonal, and it also groups data
within cluster. Such intuition shows that CASS is more accurate to
reveal the true data structure for subspace segmentation.
\begin{algorithm}[t]
\caption{Solving Problem (\ref{CASSl2}) by ADM}
\textbf{Input:} data matrix $X$, parameter $\lambda$.   \\
\textbf{Initialize:} $w^0$, $Y^0$, $\mu^0$, $\rho$, $\mu_{max}$, $\varepsilon$, $t=0$. \\
\textbf{Output:} coefficient $w^*$. \\
\textbf{while} not converge \textbf{do}
\begin{enumerate}
  \item fix the others and update $J$ by\\
    $J^{t+1}=\arg\min \frac{\lambda}{\mu^t}||J||_*+\frac{1}{2}||J-(X\text{Diag}(w^t)-\frac{1}{\mu^t}Y^t)||_F^2$.
  \item fix the others and update $w$ by  \\
    $w^{t+1}=A(X^Ty+\text{diag}(X^T(Y^t+\mu^t J^{t+1})))$,\\
    where $A=(X^TX+\mu^t\text{Diag}(\text{diag}(X^TX)))^{-1}$.
  \item update the multiplier\\
    $Y^{t+1}=Y^t+\mu^t(J^{t+1}-X\text{Diag}(w^{t+1}))$.
  \item update the parameter by $\mu^{t+1}=\min(\rho\mu^t,\mu_{max}).$
  \item check the convergence conditions\\
    $||J^{t+1}-J^t||_\infty\leq \varepsilon$,\\
    $ ||w^{t+1}-w^{t}||_\infty\leq \varepsilon$, \\
    $||J^{t+1}-X\text{Diag}(w^{t+1})||_{\infty}\leq \varepsilon  $.
  \item $t=t+1$.\\
\textbf{end while}
\end{enumerate}
\label{Algoopti}
\end{algorithm}

\subsection{The grouping effect}
\label{sec_grouping}
It has been shown in \cite{LSR} that the effectiveness of LSR by $\ell^2$-regularization comes from the grouping effect, \emph{i.e.}, the coefficients of a group of correlated data are approximately equal. In this work, we show that trace Lasso also has the grouping effect for correlated data.

\begin{theo}\label{Thm_grouping}
Given a data vector $y\in\mathbb{R}^d$, data points $X=[x_1,\cdots,x_n]\in\mathbb{R}^{d\times n}$ and parameter $\lambda>0$. Let $w^*=[w_1^*,\cdots,w_n^*]^T\in\mathbb{R}^n$ be the optimal solution to problem (\ref{CASSl2}). If $x_i\rightarrow x_j$, then $w_i^*\rightarrow w_j^*$.
\end{theo}

The proof of the Theorem \ref{Thm_grouping} can be found in the supplementary materials.

%

If each column of $X$ is normalized, $x_i=x_j$ implies that the
sample correlation $r=x_i^Tx_j=1$. Namely $x_i$ and $x_j$ are
highly correlated. Then these two data points will be grouped
together by CASS due to the grouping effect. Illustrations of the
grouping effect are shown in Figures \ref{fig_grouping} and
\ref{fig_YaleB_graphs}. One can see that the connections within
cluster by CASS are dense, similar to LRR and LSR. The grouping
effect of CASS may be weaker than LRR and LSR, since it also
encourages sparsity between clusters, but it is sufficient enough
for grouping correlated data together.

\subsection{Optimization}

Performing CASS needs to solve the convex optimization problem (\ref{CASSl2}), which can be optimized by off-the-shelf solvers. The work in \cite{TraceLasso} introduces an iteratively reweighted least squares method for solving problem (\ref{CASSl2}), but the solution is not necessarily globally optimal due to a trick by adding a term to avoid the non-invertible issue. Motivated by the optimization method used in low-rank minimization \cite{boyd2011distributed,LADMPSAP}, we adopt the Alternating Direction Method (ADM) to solve problem (\ref{CASSl2}). We first convert it to the following equivalent problem:
\begin{equation}
    \begin{split}
        \min_{J,w} \ & \frac{1}{2}||y-Xw||_2^2+\lambda||J||_*\\
        \text{s.t.}     \ & J=X\text{Diag}(w).\\
    \end{split}
\end{equation}
This problem can be solved by the ADM method, which operates on
the following augmented Lagrangian function:
\begin{equation}\label{Lagfun}
    \begin{array}{l}
L(J,w)=\frac{1}{2}||y-Xw||_2^2+\lambda||J||_*                                   \\
+\text{tr}(Y^T(J-X\text{Diag}(w)))+\frac{\mu}{2}||J-X\text{Diag}(w)||_F^2,
    \end{array}
\end{equation}
where $Y\in\mathbb{R}^{d\times n}$ is the Lagrange multiplier and
$\mu>0$ is the penalty parameter for violation of the linear
constraint. We can see that $L(J,w)$ is separable, thus it can be
decomposed into two subproblems and minimized with regard to $J$
and $w$, respectively.
The whole procedure for solving problem (\ref{CASSl2}) is outlined
in the Algorithm \ref{Algoopti}. It iteratively solves two
subproblems which have closed form solutions. By the theory of ADM
and the convexity of problem (\ref{CASSl2}), Algorithm
\ref{Algoopti} converges globally.



\begin{algorithm}[t]
\caption{Correlation Adaptive Subspace Segmentation}
\textbf{Input:} data matrix $X$, number of subspaces $k$
\begin{enumerate}
  \item Solve problem (\ref{CASSl2}) for each data point in $X$ to obtain the coefficient matrix $W^*$, where $X$ in  (\ref{CASSl2}) should be replaces by $X_{\hat{i}}=[x_1,\cdots,x_{i-1},x_{i+1},\cdots,x_n]$.
  \item Construct the affinity matrix by $(|W^*|+|{W^*}^T|)/2$.
  \item Segment the data into $k$ groups by Normalized Cuts.
\end{enumerate}
\label{AlgoASS}
\end{algorithm}
\subsection{The segmentation algorithm}
For solving the subspace segmentation problem by trace Lasso, we
first solve problem (\ref{CASSl2}) for each data point $x_i$ with
$X_{\hat{i}}=[x_1,\cdots,x_{i-1},x_{i+1},\cdots,x_n]$ which
excludes $x_i$ itself, and obtain the corresponding coefficients.
Then these coefficients can be arranged as a matrix $W^*$. The
affinity matrix is defined as $(|W^*|+|{W^*}^T|)/2$. Finally, we
use the Normalized Cuts (NCuts) \cite{NormaCut} to segment the
data into $k$ groups. The whole procedure of CASS algorithm is
outlined in the Algorithm \ref{AlgoASS}.

\section{Experiments}
In this section, we apply CASS for subspace segmentation on three databases: the Hopkins 155 \footnote{http://www.vision.jhu.edu/data/hopkins155/} motion database, Extended Yale B database \cite{YaleBdatabase} and MNIST database \footnote{http://yann.lecun.com/exdb/mnist/} of handwritten digits. CASS is compared with SSC, LRR and LSR which are the representative and state-of-the-art methods for subspace segmentation. The derived affinity matrices from all algorithms are also evaluated for the semi-supervised learning task on the Extended Yale B database. For fair comparison with previous works, we follow the experimental settings as in \cite{LSR}. The parameters for each method are tuned to achieve the best performance. The segmentation accuracy/error is used to evaluate the subspace segmentation performance. The accuracy is calculated by the best matching rate of the predicted label and the ground truth of data \cite{SSC}.
\subsection{Data sets and experimental settings}
\label{subsec_exp} \textbf{Hopkins 155} motion database contains
156 sequences, each of which has 39$\sim$550 data points drawn
from two or three motions (a motion corresponds to a subspace).
Each sequence is a sole data set and so there are 156 subspace
segmentation problems in total. We first use PCA to project the
data into a 12-dimensional subspace. All the algorithms are
performed on each sequence, and the maximum, mean and standard
deviation of the error rates are reported.

\begin{table}[!t]
\caption{\small{The segmentation errors (\%) on the Hopkins 155 database.}}
\label{Tab_hop}
\centering
\begin{tabular}{|l  c c c c c |}
\hline
\multicolumn{6}{|c|} {Comparison under the same setting}  \\
 &  kNN & SSC & LRR & LSR & CASS     \\ \hline
MAX  & 45.59 & 39.53 &  36.36 & 36.36 & \textbf{32.85}        \\
MEAN  & 13.44 & 4.02 & 3.23 & 2.50 & \textbf{2.42}    \\
STD   & 12.90 & 10.04 & 6.60 & \textbf{5.62} & 5.84    \\ \hline\hline
\multicolumn{6}{|c|} {Comparison to state-of-the-art methods}  \\
    & &ã€€SSC & LRR & LatLRR & CASS      \\ \hline
MEAN  & & 2.18 & 1.71 & \textbf{0.85} & 1.47\\\hline
\end{tabular}
\end{table}
\begin{table}[!t]
\caption{\small{The segmentation accuracies (\%) on the Extended Yale B database.}}
\label{Tab_YaleB}
\centering
\begin{tabular}{| c| c| c| c| c|c|}
\hline
&kNN & SSC & LRR & LSR & CASS \\\hline
5 subjects & 56.88 & 80.31 & 86.56 & 92.19 & \textbf{94.03} \\ \hline
8 subjects & 52.34 & 62.90 & 78.91 & 80.66 & \textbf{91.41} \\ \hline
10 subjects & 50.94 & 52.19 & 65.00 & 73.59 & \textbf{81.88} \\ \hline
\end{tabular}
\end{table}

\textbf{Extended Yale B} is challenging for subspace segmentation
due to large noises. It consists of 2,414 frontal face images of
38 subjects under various lighting, poses and illumination
conditions. Each subject has 64 faces. We construct three subspace
segmentation tasks based on the first 5, 8 and 10 subjects face
images of this database. The data are first projected into a
$5\times 6$, $8\times 6$, and $10\times 6$-dimensional subspace by
PCA, respectively. Then the algorithms are employed on these three
tasks and the accuracies are reported.

To further evaluate the effectiveness of CASS for other learning
problems, we also use the derived affinity matrix for
semi-supervised learning. The Markov random walks algorithm
\cite{ranwalk} is employed in this experiment. It performs a
$t$-step Markov random walk on the graph or affinity matrix. The
influence of one example to another example is proportional to the
affinity between them. We test on the 10 subjects face
classification problem. For each subject, 4, 8, 16 and 32 face
images are randomly selected to form the training data set, and
the remaining for testing. Our goal is to predict the labels of
the test data by Markov random walks \cite{ranwalk} on the
affinity matrices learnt by $k$NN, SSC, LRR, LSR and CASS. We
experimentally select $k=6$ neighbors. The experiment is repeated
for 20 times, and the accuracy and standard deviation are reported
for evaluation.

\textbf{MNIST} database of handwritten digits is also widely used
in subspace learning and clustering \cite{mnistref}. It has 10
subjects, corresponding to 10 handwritten digits, 0$\sim$9. We
select a subset with a similar size as in the above face
clustersing problem for this experiment, which consists of the
first 50 samples of each subject. The accuracies of SSC, LRR, LSR
and CASS are reported.

\begin{figure}[!t]
\centering
\includegraphics[width=0.45\textwidth]{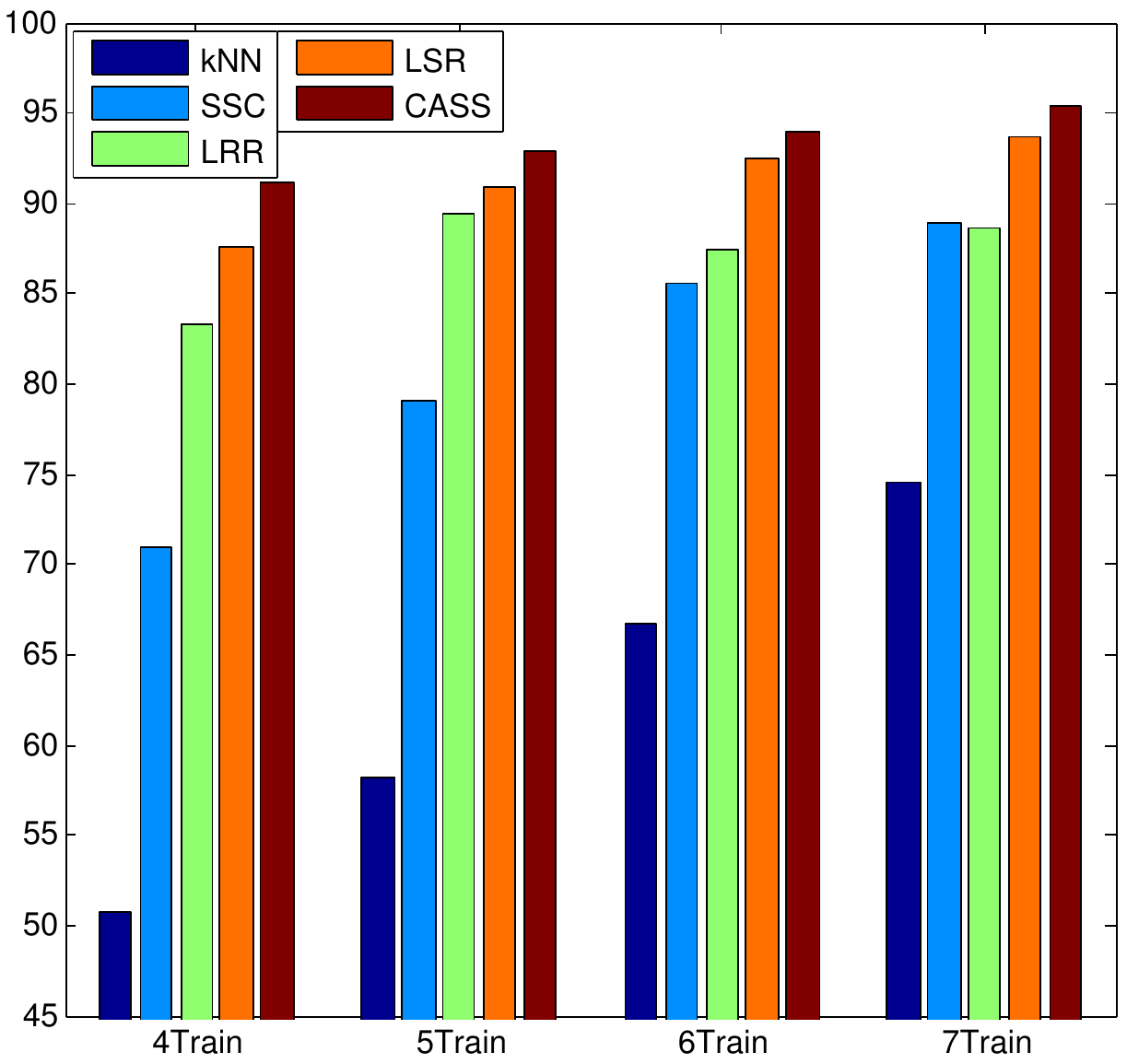}
\caption{\small{Comparison of classification accuracy (\%) and
standard deviation of different semi-supervised learning based on
different affinity matrices on the Extended Yale B (10 subjects)
database.}} \label{fig_semi}
\end{figure}

\subsection{Experimental results}
Table \ref{Tab_hop} tabulates the motion segmentation errors of
four methods on the Hopkins 155 database. It shows that CASS gets
a misclassification error of 2.42$\%$ for all 156 sequences, while
the best previously reported result is 2.50$\%$ by LSR. The
improvement of CASS on this database is limited due to many
reasons. First, previous methods have performed very well on the
data with only slight corruptions, and thus the room for
improvement is limited. Second, the reported error is the mean of
156 segmentation errors, most of which are zeros. So even if there
are some high improvements on some challenging sequences, the
improvement of the mean error is also limited. Third,  the
correlation of data is strong as the dimension of each affine
subspace is no more than three \cite{SSC} \cite{LSR}, thus CASS
tends to be close to LSR in this case. Due to the dimensionality
reduction by PCA and sufficient data sampling in each motion, CASS
may behave like LSR with a strong grouping effect. Furthermore, in
order to compare with the state-of-the-art methods, we follow the
post-processing in \cite{LRRpami}, which may \emph{not} be optimal
for CASS, and the error of CASS is reduced to 1.47$\%$. But the
best performance by Latent LRR \cite{liu2011latent} is 0.85$\%$.
It is much better than other methods. That is because Latent LRR
further employs unobserved hidden data as the dictionary and
\emph{has complex pre-processing and post-processing with several
parameters}. The idea of incorporating unobserved hidden data may
also be considered in CASS. This will be our future work.

\begin{figure}[!t]
\centering
\includegraphics[width=0.5\textwidth]{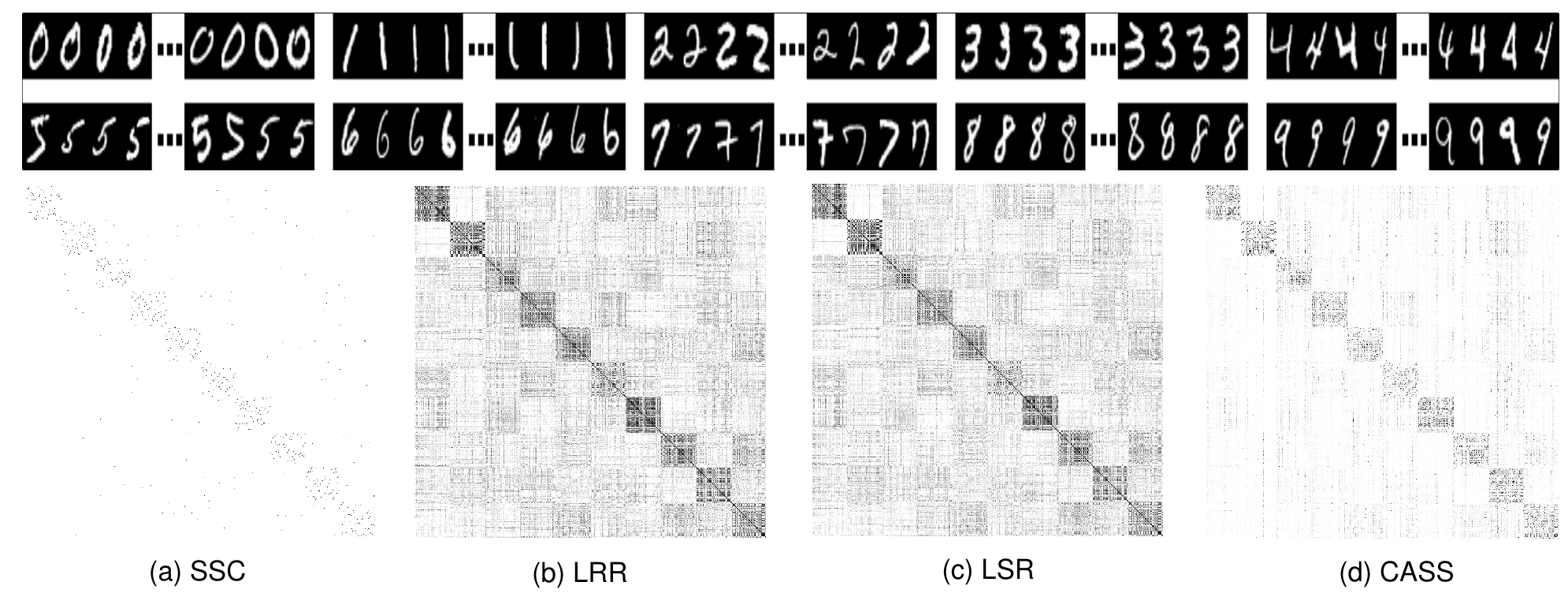}
\caption{\small{The affinity matrices derived by (a) SSC, (b) LRR, (c) LSR, and (d) CASS on the MNIST database.}}
\label{fig_digit_graphs}
\end{figure}
\begin{table}[!t]
\caption{\small{The segmentation accuracies (\%) on the MNIST database.}}
\label{Tab_MNIST}
\centering
\begin{tabular}{| c|c| c| c| c|c|c|}
\hline
&kNN & SSC & LRR & LSR & CASS \\ \hline
ACC. &61.00  & 62.60 &  66.80 & 68.00 & \textbf{73.80} \\ \hline
\end{tabular}
\end{table}
Table \ref{Tab_YaleB} shows the clustering result on the Extended
Yale B database. We can see that CASS outperforms SSC, LRR and LSR
on all these three clustering tasks. In particular, CASS gets
accuracies of 94.03$\%$, 91.41$\%$, and 81.88$\%$ for face
clustering with 5, 8, and 10 subjects, respectively, which
outperforms the state-of-the-art method LSR. For the 5 subjects
face clustering problem, all these four methods perform well, and
no big improvement is made by CASS. But for the 8 subjects and 10
subjects face clustering problems, CASS achieves significant
improvements. For these two clustering tasks, both LRR and LSR
perform much better than SSC, which can be attributed to the
strong grouping effect of the two methods. However, both the two
methods lack the ability of subset selection, and therefore may
group some data points between clusters together. CASS not only
preserves the grouping effect within cluster but also enhances the
sparsity between clusters. The intuitive comparison of these four
methods can be found in Figure \ref{fig_YaleB_graphs}. It confirms
that CASS usually leads to an approximately block diagonal
affinity matrix which results in a more accurate segmentation
result. This phenomenon is also consistent with the analysis in
Theorems \ref{ThmBlock} and \ref{Thm_grouping}.

For semi-supervised learning, the comparison of the classification
accuracies is shown in Figure \ref{fig_semi} with different
numbers of training data. CASS achieves the best performance and
the accuracies on these settings are all above $90\%$. Notice that
they are much higher than the clustering accuracies in Table
\ref{Tab_YaleB}. This is mainly due to the mechanism of
semi-supervised learning which makes use of both labeled and
unlabeled data for training. The accurate graph construction is
the key step for semi-supervised learning. This example shows that
the affinity matrix by trace Lasso is also effective for
semi-supervised learning.

Table \ref{Tab_MNIST} shows the clustering accuracies by SSC, LRR,
LSR, and CASS on the MNIST database. The comparison of the derived
affinity matrices by these four methods is illustrated in Figure
\ref{fig_digit_graphs}. We can see that CASS obtains an affinity
matrix which is close to block diagonal by preserving the grouping
effect. None of these four methods performs perfectly on this
database. Nonetheless, our proposed CASS method achieves the best
accuracy $73.80\%$. The main reason may lie in the fact that the
handwritten digit data do not fit the subspace structure well.
This is also the main challenge for real-world applications by
subspace segmentation.

\section{Conclusions and Future Work}
In this work, we propose the Correlation Adaptive Subspace
Segmentation (CASS) method by using the trace Lasso. Compared with
the existing SSC, LRR, and LSR, CASS simultaneously encourages
grouping effect and sparsity. The adaptive advantage of CASS comes
from the mechanism of trace Lasso which balances between
$\ell^1$-norm and $\ell^2$-norm. In theory, we show that CASS is
able to reveal the true segmentation result when the subspaces are
independent. The grouping effect of trace Lasso is firstly
established in this work. At last, the experimental results on the
Hopkins 155, Extended Yale B, and MNIST databases show the
effectiveness of CASS. Similar improvement can also be observed in
semi-supervised learning setting on the Extended Yaled B database.
However, there still remain many problems for future exploration.
First, the data itself, which may be noisy, are used as the
dictionary for linear construction. It may be better to learn a
compact and discriminative dictionary for trace Lasso. Second,
trace Lasso may have many other applications, \emph{i.e.}
classification, dimensionality reduction, and semi-supervised
learning. Third, more scalable optimization algorithms should be
developed for large scale subspace segmentation.
\section*{Acknowledgements}
This research is supported by the Singapore National Research Foundation under its International Research Centre @Singapore Funding Initiative and administered by the IDM Programme Office. Z. Lin is supported by National Natural Science Foundation of China (Grant nos. 61272341, 61231002, and 61121002).

{
\small
\bibliographystyle{ieee}
\bibliography{CASSreference}
}

\end{document}